# PHÂN CỤM MỜ VỚI TRỌNG SỐ MŨ NGÔN NGỮ

**Lê Thái Hưng[1], Trần Đình Khang[1], Lê Văn Hưng[3]**

[1] Trường Đại học Bách khoa Hà Nội
[2] Trường Đại học Mỏ Địa chất

*leehung314@gmail.com, khangtd@soict.hust.edu.vn*

***TÓM TẮT*** - *Bài báo này được thực hiện nhằm mục đích nghiên cứu tìm hiểu thuật toán phân cụm FCM và các ý tưởng cải tiến đã có; tiến hành phân tích và phát hiện những đặc điểm phù hợp trong thuật toán FCM có thể áp dụng được đại số gia tử - một lý thuyết sử dụng đại số trong việc biểu diễn giá trị của các biến ngôn ngữ. Từ đó, đề xuất một hướng cải tiến mới, đó là sử dụng lý thuyết đại số gia tử vào trọng số mũ của thuật toán FCM và sau cùng là xây dựng cài đặt một thuật toán phân cụm mờ sử dụng đại số gia tử để có thể áp dụng giải quyết bài toán phân cụm trong các ứng dụng thực tế.*

***Từ khóa*** - *Phân cụm mờ, Thuật toán FCM, Đại số gia tử, Thuật toán HAmFCM.*

## I. ĐẶT VẤN ĐỀ

Bài toán phân cụm là bài toán phân hoạch các đối tượng vào các nhóm sao cho những đối tượng cùng một nhóm thì có nhiều điểm giống nhau trong khi những đối tượng khác nhóm thì có ít điểm giống nhau. Đây là một bài toán có ý nghĩa quan trọng trong nhiều lĩnh vực của đời sống con người, giúp chúng ta hiểu rõ hơn về bản chất, cấu trúc của dữ liệu để từ đó xử lý dữ liệu được hiệu quả hơn. Nhằm giải quyết bài toán này, đã có rất nhiều thuật toán phân cụm được đề xuất, trong đó có thể nói tiêu biểu nhất là thuật toán FCM (Fuzzy c-Means) [1] ra đời vào năm 1981. Kể từ đó đến nay, đã có rất nhiều các phương pháp mới được đưa ra dựa trên nền tảng FCM, nhằm cải tiến, khắc phục các điểm còn hạn chế, giúp cải thiện hơn nữa khả năng phân cụm của thuật toán này trong nhiều trường hợp khác nhau.

Bài báo được thực hiện nhằm mục đích nghiên cứu tìm hiểu thuật toán phân cụm FCM và các ý tưởng cải tiến đã có; tiến hành phân tích và phát hiện những đặc điểm phù hợp trong thuật toán FCM có thể áp dụng được đại số gia tử (Hedge Algebra - HA) - một lý thuyết sử dụng đại số trong việc biểu diễn giá trị của các biến ngôn ngữ [3,4,5,6]. Từ đó, bài báo đề xuất một hướng cải tiến mới, đó là sử dụng lý thuyết đại số gia tử vào trọng số mũ của thuật toán FCM, và sau cùng là xây dựng cài đặt một thuật toán phân cụm mờ sử dụng đại số gia tử HAmFCM để có thể áp dụng giải quyết bài toán phân cụm trong các ứng dụng thực tế. Phần thực nghiệm với các bộ dữ liệu UCI [7] có so sánh với các phương pháp cải tiến khác FCMT2I, FCMT2G [2,8] để minh hoạ hiệu quả của phương pháp.

Cấu trúc của bài báo như sau, Phần II trình bày về phân cụm mờ và thuật toán FCM, Phần III là đề xuất cải tiến HAmFCM và Phần IV là các thực nghiệm.

## II. BÀI TOÁN PHÂN CỤM MỜ

***A. Bài toán phân cụm:*** là bài toán phân hoạch các đối tượng vào các nhóm sao cho những đối tượng cùng một nhóm thì có nhiều điểm giống nhau trong khi những đối tượng khác nhóm thì có ít điểm giống nhau. Đây là một bài toán có ý nghĩa quan trọng trong nhiều lĩnh vực của đời sống con người, giúp chúng ta hiểu rõ hơn về bản chất, cấu trúc của dữ liệu để từ đó xử lý dữ liệu được hiệu quả hơn. Nhằm giải quyết bài toán này, đã có rất nhiều thuật toán phân cụm được đề xuất, trong đó có thể nói tiêu biểu nhất là thuật toán FCM (Fuzzy c-Means) ra đời vào năm 1981. Kể từ đó đến nay, đã có rất nhiều các phương pháp mới được đưa ra dựa trên nền tảng FCM, nhằm cải tiến, khắc phục các điểm còn hạn chế, giúp cải thiện hơn nữa khả năng phân cụm của thuật toán này trong nhiều trường hợp khác nhau.

***B. Thuật toán FCM***

Trước hết, ta sẽ định nghĩa một cách chính xác bài toán phân cụm. Xét bài toán phân cụm n phần tử trong tập dữ liệu:

$$X = \{x_1, x_2, ..., x_n\}$$

Mỗi phần tử $x_i \in X$, i=1, 2,…n là một vector d chiều. Ta định nghĩa một c-phân cụm của X là một phân hoạch X vào c tập (cụm) $C_1, C_2, ...C_c$, sao cho 3 điều kiện sau được thỏa mãn:

- $C_i \neq \emptyset$ với $i = 1, 2, ..., c$
- $\bigcup_{i=1}^{c} C_i = X$
- $C_i \cap C_j = \emptyset$ với $i \neq j$; $i, j = 1, 2, ..., c$

Trong thuật toán FCM, hàm thuộc được biểu diễn rời rạc thành một ma trận thực U kích thước n × c:
- $U(i, k)$ là độ thuộc của phần tử $x_i$ vào cụm thứ $C_k$: $1 \leq i \leq n$; $1 \leq k \leq c$
- U có $n \times c$ phần tử
- $0 \leq U(i, k) \leq 1$

- $U(i,k)$ càng lớn thì phần tử $x_i$ càng có nhiều khả năng thuộc vào cụm $k$
- $U$ được gọi là ma trận thuộc

Dựa trên mô hình ma trận thuộc này, một hàm mục tiêu được xác định sao cho thuật toán phân cụm phải tối thiểu hóa hàm mục tiêu. Thuật toán FCM sử dụng hàm mục tiêu là:

$$J(U,C) = \sum_{i=1}^{n}\sum_{k=1}^{c} U(i,k)^m \|X(i) - C(k)\|^2 \quad (1)$$

Ở đây đã sử dụng các ký hiệu sau:

- $X(i)$ là vector giá trị của phần tử $x_i$
- $C$ là tập vector giá trị tâm cụm của $c$ cụm $C_1, C_2, ... C_c$
- $C(k)$ là vector giá trị tâm cụm $C_k$
- $D(i,k) = \|X(i) - C(k)\|^2$ là một độ đo khoảng cách giữa 2 vector $X(i)$ và $C(k)$. Thuật toán FCM nguyên bản sử dụng độ đo Euclid
- $m$ là tham số của thuật toán, gọi là trọng số mũ

**Thuật toán chi tiết:**

- ❖ Bước 1: Khởi tạo giá trị $U, C$, chọn giá trị tham số $m$ ($m > 1$ và thường được chọn trong khoảng [2,40])
- ❖ Bước 2: Với mọi $k$, cập nhật giá trị $C$ theo công thức:

$$C(k) = \frac{\sum_{i=1}^{n} U(i,k)^m X(i)}{\sum_{i=1}^{n} U(i,k)^m} \quad (2)$$

- ❖ Bước 3: Với mọi $i, k$ cập nhật giá trị $U$ theo công thức:

$$U(i,k) = \left(\sum_{j=1}^{c}\left(\frac{D(i,k)}{D(i,j)}\right)^{\frac{2}{m-1}}\right)^{-1} \quad (3)$$

Với $D(i,k) = \|X(i) - C(k)\|^2$

- ❖ Bước 4: Kiểm tra điều kiện dừng: Tính giá trị hàm mục tiêu $J$ theo công thức (1.1). Nếu giá trị $J$ thay đổi, quay lại bước 2, trái lại ta kết thúc thuật toán.

Ta có nhận xét rằng, thuật toán FCM gồm 2 bước chính là cập nhật tâm cụm-công thức (1.2) và cập nhật ma trận thuộc-công thức (1.3). Hai bước này được thực hiện luân phiên nhằm tối thiểu hóa hàm $J$. Thuật toán sẽ dừng khi $J$ hội tụ. Sau khi kết thúc thuật toán, ta sẽ ra quyết định phân cụm dựa vào giá trị $U$.

**Nhược điểm thuật toán FCM:**

FCM là một thuật toán đột phá so với các thuật toán phân cụm rõ khác vì FCM có khả năng biểu diễn tổng quát hơn và cho kết quả phân cụm chính xác hơn trong nhiều trường hợp. Nhưng FCM vẫn còn đó những nhược điểm:

- Thuật toán ngầm định mỗi cụm có một tâm cụm và các phần tử thuộc cụm nào thì nằm gần tâm cụm đó
- Kết quả của thuật toán phụ thuộc vào khởi tạo U, C ban đầu. Nếu khởi tạo không tốt, có thể bị hội tụ địa phương
- Thuật toán nhạy cảm với sự xuất hiện của nhiễu và các phần tử ngoại lai do chưa có mô hình biểu diễn nhiễu
- Phân cụm chưa xác đáng với các đối tượng nằm ở ranh giới giữa các cụm
- Chưa có tiêu chí cụ thể để lựa chọn giá trị cho tham số m, thường chọn bằng thử nghiệm nhiều lần

Với các nhược điểm đó, FCM vẫn cần được nghiên cứu để được cải thiện hơn nữa. Bài báo này sẽ đề xuất một hướng cải tiến cho FCM bằng cách sử dụng trọng số mũ ngôn ngữ xác định bởi đại số gia tử.

*C. Ý tưởng cải tiến FCM*

Khi đưa ra quyết định phân cụm, FCM dựa vào công thức:

$$U(i,k) = \left(\sum_{j=1}^{c}\left(\frac{D(i,k)}{D(i,j)}\right)^{\frac{2}{m-1}}\right)^{-1}$$

Công thức này có đặc điểm, khi m càng nhỏ, độ tuyệt đối của U (độ dốc của đồ thị) càng cao, như hình sau minh họa trong trường hợp $c = 2$:

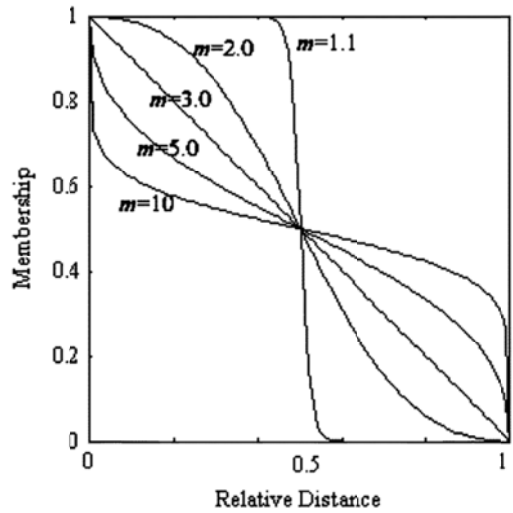

Rõ ràng, với các phần tử ở ranh giới cụm (khoảng cách tương đối lớn), độ tuyệt đối của U phải thấp hơn so với các phần tử ở gần tâm cụm (khoảng cách tương đối nhỏ). Lý do là bởi vì ta chưa thể đưa ra quyết định phân cụm với các phần tử ở ranh giới ngay được, cho nên ta cần đồ thị U thoải hơn để linh động hơn trong việc chọn cụm cho các phần tử này. Trong khi đó, các phần tử gần tâm cụm nên có đồ thị U dốc hơn để đảm bảo kiểm soát sự phân cụm cho các phần tử đó luôn rơi vào cụm gần nhất. **Do FCM chỉ sử dụng một giá trị m duy nhất, nên mức độ tuyệt đối của U là như nhau với các phần tử gần tâm cụm và các phần tử ở ranh giới, điều này là chưa hợp lý.**

Để hợp lý, ta có quy luật sau:

- Khi khoảng cách tương đối nhỏ (phần tử gần như đã ở rất gần một cụm nào đó) thì m nên nhỏ, độ tuyệt đối của U tăng lên
- Khi khoảng cách tương đối lớn (phần tử nằm ở ranh giới giữa các cụm hoặc chưa quá gần một cụm nào cả) thì m nên lớn, phản ảnh độ tuyệt đối của U giảm xuống

Tức là m thay đổi theo khoảng cách tương đối, như vậy để biểu diễn $m$, ta cần sử dụng một tập các giá trị $M(i,k)$ nằm trong khoảng [m_min,m_max] và các công thức cập nhật sẽ trở thành:

$$U(i,k) = \left( \sum_{j=1}^{c} \left( \frac{D(i,k)}{D(i,j)} \right)^{\frac{2}{M(i,k)-1}} \right)^{-1} \tag{4}$$

$$C(k) = \frac{\sum_{i=1}^{n} U(i,k)^{M(i,k)} X(i)}{\sum_{i=1}^{n} U(i,k)^{M(i,k)}} \tag{5}$$

Hàm mục tiêu trở thành:

$$J(U,C) = \sum_{i=1}^{n} \sum_{k=1}^{c} U(i,k)^{M(i,k)} \|X(i) - C(k)\|^2 \tag{6}$$

Vấn đề bây giờ là xây dựng tập giá trị M(i,k) như thế nào để thỏa mãn quy luật nói ở trên.

### III. THUẬT TOÁN HAmFCM

#### A. Tập mờ trọng số mũ

Đầu tiên, ta cần định nghĩa lại khoảng cách tương đối:

$$T_{i,k} = \frac{D(i,k)}{\sum_{j=1}^{c} D(i,j)} \tag{7}$$

Dễ thấy độ đo này có tính chất:
- Khi $x_i$ tiến rất gần $C_k$, $D(i,k)$ tiến về 0, $T_{i,k}$ sẽ có giá trị tiến về 0, nhưng luôn lớn hơn 0
- Khi $x_i$ tiến ra xa $C_k$, $D(i,k)$ tiến ra $\infty$, $T_{i,k}$ sẽ luôn lớn hơn 0 và có xu hướng tăng dần

Ngoài ra, mẫu của $T_{i,k}$ luôn khác 0, nên an toàn trong quá trình tính toán. Thêm nữa, để phù hợp với miền định lượng ngữ nghĩa của HA sau này (từ 0 đến 1), ta tiến hành chuẩn hóa $T_{i,k}$ về miền [0,1] theo công thức như sau:

$$Q_{i,k} = \frac{T_{i,k} - T_{min}}{T_{max} - T_{min}} \tag{8}$$

Tiếp theo, nhắc lại quy luật cần thỏa mãn:
- $Q_{i,k}$ nhỏ thì $M(i,k)$ nhỏ
- $Q_{i,k}$ lớn thì $M(i,k)$ lớn

Một hàm ánh xạ tuyến tính đơn giản có thể giúp tính $M(i,k)$ từ $Q_{i,k}$ như sau:

$$M(i,k) = Q_{i,k} \times (m\_max - m\_min) + m\_min \tag{9}$$

Như vậy, ta đã có phương pháp để xác định giá trị các trọng số mũ trong ma trận trọng số mũ $M$. Ở đây, công thức (9) ngầm định mối quan hệ giữa khoảng cách tương đối và trọng số mũ là quan hệ tuyến tính cùng tăng cùng giảm. Điều này tất nhiên là chưa phản ánh chính xác thực tế, do đó, ta cần gắn với mỗi giá trị trọng số mũ một độ đo thể hiện độ tin cậy. Độ tin cậy, kí hiệu $R(i,k)$ thể hiện mức độ tin cậy của giá trị $M(i,k)$ khi tính theo công thức (9). Độ tin cậy có tính chất:
- $0 \leq R(i,k) \leq 1$
- $R(i,k)$ nhỏ, độ chắc chắn của giá trị $M(i,k)$ nhỏ, nhiều khả năng sẽ phải thay đổi bằng một giá trị khác phù hợp hơn
- $R(i,k)$ lớn, độ chắc chắn của giá trị $M(i,k)$ lớn, nhiều khả năng sẽ phải giữ nguyên

Cách tính độ tin cậy $R(i,k)$:

$$R(i,k) = R(M(i,k)) = R(Q_{i,k}) = f_m\left(v^{-1}(Q_{i,k})\right) \tag{10}$$

Từ đây, ta có **tập mờ trọng số mũ** tương ứng là:

$$\overline{A} = \sum_{\substack{\forall M(i,k) \in \\ [m\_min, m\_max]}} \frac{R(i,k)}{M(i,k)} \tag{11}$$

*B. Cập nhật trọng số mũ*

Nếu sử dụng trực tiếp $M(i,k)$ tính theo công thức (9) thì hơi cứng nhắc vì quan hệ giữa khoảng cách tương đối và trọng số mũ có thể không phải là quan hệ tuyến tính. Do đó để làm linh hoạt, ta sẽ dùng độ tin cậy của trọng số mũ để cập nhật lại giá trị trọng số mũ. Cách làm như sau, tại vòng lặp thứ t, tính $Q_{i,k}^{(*)}$ từ tập dữ liệu, sau đó cập nhật giá trị $Q_{i,k}^{(t)}, M(i,k)^{(t)}$ như sau:

$$Q_{i,k}^{(t)} = Q_{i,k}^{(t-1)} + R(i,k) \times \left(Q_{i,k}^{(*)} - Q_{i,k}^{(t-1)}\right)$$
$$M(i,k)^{(t)} = Q_{i,k}^{(t)} \times (m\_max - m\_min) + m\_min \tag{12}$$

Ở đây dễ thấy, nếu độ tin cậy là tuyệt đối $R(i,k) = 1$, khi đó $Q_{i,k}^{(t)} = Q_{i,k}^{(*)}$ và $M(i,k)$ được tính y hệt như (9). Và khi độ tin cậy $R(i,k) = 0$, $Q_{i,k}^{(t)} = Q_{i,k}^{(t-1)}$, tức là không có sự cập nhật giá trị $Q_{i,k}^{(t)}$ và $M(i,k)^{(t)}$. Nói chung, độ tin cậy càng cao, giá trị $M(i,k)$ càng tiến gần đến giá trị tính được bằng công thức (9)-tức là tiến gần đến một giá trị mới xác định từ ánh xạ tuyến tính của trạng thái phân cụm hiện tại, còn khi độ tin cậy thấp, giá trị $Q_{i,k}^{(t)}$ và $M(i,k)^{(t)}$ có xu hướng ít tuân theo quy luật tuyến tính của sự thay đổi của trạng thái phân cụm.

*C. Tối ưu tham số HA*

Trong công thức (3), ảnh hưởng của đại lượng độ tin cậy là rất to lớn, quyết định trực tiếp đến giá trị $M(i,k)$ và độ tin cậy này hoàn toàn xác định bởi HA. Nói cách khác, giá trị các tham số mờ của HA cũng tham gia vào quá trình xác định $M(i,k)$. Việc cập nhật thay đổi các tham số HA làm thay đổi khả năng biểu diễn ngôn ngữ của HA, từ đó tác động đến hiệu quả của thuật toán HAmFCM. Ở đây, ta đưa ra tiêu chí cập nhật tham số HA đó là làm giảm lỗi ánh xạ ngữ nghĩa và hàm mục tiêu. Trước hết, ta định nghĩa lỗi ánh xạ ngữ nghĩa của giá trị thực $q$ chính là sai khác giữa $q$ và định lượng ngữ nghĩa của HA:

$$e(q) = |v(v^{-1}(q)) - q| \tag{13}$$

Ta thấy rằng khi lỗi này càng nhỏ, HA càng biểu diễn tốt các giá trị trọng số mũ vì các giá trị ngôn ngữ của HA có định lượng ngữ nghĩa rất gần với các giá trị $q$ - tương ứng với các trọng số mũ. Việc tìm ra một phương án cập nhật tham số HA sao cho chắc chắn làm giảm lỗi này là không dễ. Do đó, NVĐA đã sử dụng một ý tưởng heuristic như sau: khi lỗi này nhỏ, ta mong muốn HA sẽ được cập nhật ít đi, không cần thay đổi quá nhiều. Ngược lại, khi lỗi này còn lớn, các tham số của HA cần phải được cập nhật với một lượng lớn hơn với hi vọng sẽ đạt được trạng thái giúp lỗi ánh xạ ngữ nghĩa nhỏ hơn.

Thêm nữa, các tham số HA ta cần cập nhật bao gồm:
- $\mu(less), \mu(possibly), \mu(more), \mu(very)$
- $f_m(small), f_m(big)$

Sẽ là hợp lý nếu ta chỉ cập nhật các tham số có liên quan đến lỗi $e(q)$. Nói cách khác, với một giá trị $q$, ta tìm được giá trị ngôn ngữ $\hat{x} = v^{-1}(q)$, $\hat{x}$ sẽ chứa các gia tử và một phần tử sinh. Ta sẽ chỉ cập nhật $\mu, f_m$ của các gia tử và phần tử sinh này. Lấy ví dụ, nếu $\hat{x} =$ "very very more big" và $e(q)$=0.01, ta tiến hành cập nhật như sau:

$$\mu(very) = \mu(very) + e(q) * 2 = \mu(very) + 0.02$$
$$\mu(more) = \mu(more) + e(q) \quad = \mu(more) + 0.01$$
$$f_m(big) = f_m(big) + e(q) \quad = f_m(big) + 0.01 \quad (14)$$

Cách cập nhật như vậy đảm bảo nguyên tắc:
- Khi lỗi ánh xạ ngữ nghĩa nhỏ, thì lượng cập nhật là nhỏ
- Chỉ cập nhật các tham số có liên quan đến lỗi

Tuy nhiên, do đây chỉ là đề xuất mang tính heuristic, để đảm bảo tiêu chí hàm mục tiêu luôn giảm, ta phải thực hiện các phương pháp sau:
- Sau khi cập nhật tham số HA, ta tính hàm mục tiêu
- Nếu hàm mục tiêu giảm so với trước đó, tiếp tục cập nhật HA, nếu không ta dừng việc cập nhật và khôi phục trạng thái trước đó
- Ngoài ra, để đảm bảo tính đúng đắn của tham số HA (các tham số HA nằm trong khoảng từ 0 đến 1) ta còn phải chuẩn hóa tham số HA

Thực hiện các quy tắc trên không những đảm bảo thuật toán đúng về mặt toán học (làm hàm mục tiêu giảm, dẫn đến một trạng thái phân cụm tốt) mà còn đảm bảo sự hội tụ của thuật toán. Trong thực tế thử nghiệm, thường thì chỉ sau khoảng 15-20 vòng lặp, ta có thể dừng việc cập nhật HA.

### D. Thuật toán HAmFCM

Các bước chính thuật toán

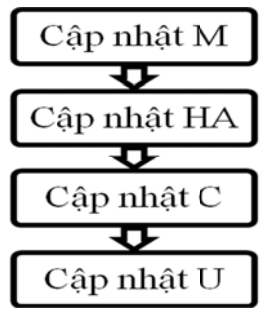

**Thuật toán** $HAmFCM$

**Đầu vào:** Tập các phần tử cần phân cụm $X$, với $X(i), i = 1,2 \ldots n$ là phần tử thứ $i$. Số cụm $c$ cần phân loại. Dải giá trị trọng số mũ $[m\_min, m\_max]$

**Đầu ra:** Vị trí tâm các cụm $C$, với $C(k), k = 1,2 \ldots c$ là giá trị vector của tâm cụm thứ $k$. Nhãn tên cụm của các phần tử, với $label(i), i = 1,2, \ldots n$ là nhãn của phần tử thứ $i$.

**Bước 1:** Khởi tạo
$HA = (AX, \{small, big\}, \{less, possibly, more, very\}, \leq)$, quan hệ $SIG$, $L = 3$;
Khởi tạo ngẫu nhiên $C$;
Khởi tạo biến $C_{old} = 0$;
Khởi tạo ma trận $U(i,k) = 0 \; \forall i,k$;
Khởi tạo ma trận $Q(i,k) = 0 \; \forall i,k$;
Khởi tạo ma trận $R(i,k) = 0 \; \forall i,k$;
Khởi tạo ma trận $M(i,k) = m_{min} \; \forall i,k$;
Khởi tạo biến $J_{old} = \infty, J = 1$;
Khởi tạo biến $update = true$;

**Bước 2:** Cập nhật $M$
FOR $i = 1$ TO $n$

    FOR $k = 1$ TO $c$
      Thực hiện (7);
      Thực hiện (8);
      Thực hiện (10);
      Thực hiện (12);
**Bước 3:** Kiểm tra hàm mục tiêu
Thực hiện (6);
IF $J > J_{old}$ THEN
    $update = false$;
$J_{old} = J$;
**Bước 4:** Cập nhật HA
IF $update$ THEN
    FOR $i = 1$ TO $n$
      FOR $k = 1$ TO $c$
        Cập nhật HA như (14)
    Chuẩn hóa tham số HA
**Bước 5:** Cập nhật $C, U$
FOR $i = 1$ TO $n$
    FOR $k = 1$ TO $c$
      Thực hiện (5);
      Thực hiện (6);
**Bước 6:** Kiểm tra điều kiện dừng
IF $C_{old} \neq C$ THEN
    $C_{old} = C$;
    Quay lại Bước 2;
ELSE
    Chuyển đến Bước 7;
**Bước 7:** Đọc kết quả phân cụm
FOR $i = 1$ TO $n$
    $label(i) = \mathrm{argmax}_k\big(U(i,k)\big)$;
RETURN $[C, label]$;

**Độ phức tạp thuật toán:**

  Ở đây, ta đánh giá độ phức tạp theo số phần tử đầu vào cần phân cụm $n$. Các yếu tố khác tham gia vào độ phức tạp của thuật toán như số cụm $c$ cần phân hoạch và các tham số của HA như $|H|, |G|$, và $L$ đều được coi là hằng số. Trước hết, ta đánh giá độ phức tạp của một vòng lặp của thuật toán này (từ bước 2 đến bước 6) như sau:

- Bước 2: Do các tính toán trong vòng lặp đều có độ phức tạp O(1), độ phức tạp cho bước này là số vòng lặp, tức là $O(n \times c) = O(n)$
- Bước 3: Độ phức tạp O(1)
- Bước 4: Do các tính toán trong vòng lặp đều có độ phức tạp O(1), độ phức tạp cho bước này là số vòng lặp, tức là $O(n \times c) = O(n)$
- Bước 5: Do các tính toán trong vòng lặp đều có độ phức tạp O(1), độ phức tạp cho bước này là số vòng lặp, tức là $O(n \times c) = O(n)$
- Bước 6: Độ phức tạp O(1)

  Ngoài vòng lặp chính này, ta còn phải xét:
- Bước 1 (khởi tạo): khởi tạo các ma trận cũng chỉ mất $O(n)$.
- Bước 7 (đọc kết quả): Tìm giá trị lớn nhất của độ thuộc cũng chỉ có độ phức tạp là $O(n)$

  Do đó, độ phức tạp của cả thuật toán sẽ là:

$$\text{số vòng lặp} \times O(n)$$

  Nhắc lại rằng thuật toán HAmFCM (và cả thuật toán FCM) chỉ dừng khi không còn sự cập nhật ở tâm cụm, tức là số vòng lặp là không biết trước, thậm chí trong trường hợp xấu, số vòng lặp có thể tiến đến vô cùng. Do đó, việc đánh giá độ phức tạp chỉ có ý nghĩa trong một vòng lặp. Và khi so sánh độ phức tạp của HAmFCM với FCM, ta cũng chỉ cần so sánh độ phức tạp trong một vòng lặp mà thôi. Và như ta thấy, dù HAmFCM có thêm 2 giai đoạn là cập nhật $M$ và cập nhật HA, thì so với FCM, độ phức tạp lý thuyết trong một vòng lặp của chúng vẫn bằng nhau và bằng $O(n)$.

**Ưu điểm của HAmFCM so với FCM:**

Như ta đã biết, điểm khác biệt lớn nhất giữa HAmFCM và FCM nằm ở chỗ, HAmFCM sử dụng một tập các giá trị trọng số mũ trong quá trình cập nhật độ thuộc và tâm cụm thay vì chỉ sử dụng một giá trị duy nhất như của FCM. Hệ quả là:

- Độ thuộc U được xác định một cách linh hoạt hơn, thay đổi theo sự thay đổi độ lớn của trọng số mũ và khoảng cách tương đối tương ứng được sử dụng. Điều này giúp quyết định phân cụm trở nên đa dạng, xác đáng hơn cho các phần tử nằm ở ranh giới giữa các cụm-những đối tượng cần sử dụng trọng số mũ khác với những phần tử nằm gần tâm cụm.

- Việc sử dụng một dải các giá trị trọng số mũ giúp đơn giản hóa công đoạn dò tìm tham số trọng số mũ. Nếu như trong FCM, ta cần phải thử chọn tham số trọng số mũ nhiều lần cho đến khi tìm được giá trị có kết quả phân cụm tốt nhất, thì HAmFCM chỉ cần xác định giá trị [m_min, m_max] một lần duy nhất và áp dụng cho hầu hết mọi bài toán. Thuật toán HAmFCM sẽ tự động cập nhật tìm ra các trọng số mũ phù hợp để áp dụng cho mỗi phần tử trong quá trình phân cụm.

Việc HAmFCM biểu diễn trọng số mũ như một tập mờ ngôn ngữ còn có vai trò giúp chúng ta hiểu rõ hơn ý nghĩa của giá trị các trọng số mũ trong quá trình vận hành thuật toán. Mô tả các con số thành những khái niệm ngôn ngữ của con người để dễ hình dung, dễ nắm bắt và điều chỉnh là đặc tính của đại số gia tử. HAmFCM đã kế thừa được tính chất này.

## IV. THỰC NGHIỆM

Dưới đây là hình ảnh minh họa giao diện chương trình:

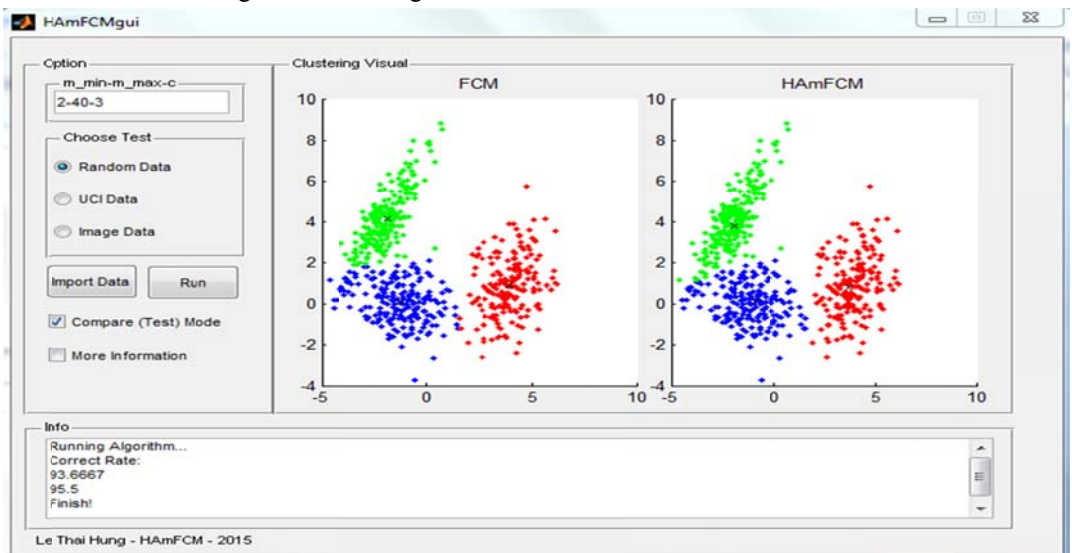

**Hình 1.** Giao diện chương trình

Chương trình cho phép người dùng nhập tham số của thuật toán HAmFCM, đó là dải giá trị trọng số mũ [m_min, m_max] và số cụm cần phân hoạch c. Giao diện bên trái là lựa chọn các chế độ thử nghiệm thuật toán. Giao diện bên phải là minh họa hình ảnh diễn biến quá trình phân cụm các phần tử của bộ dữ liệu. Giao diện phía dưới hiển thị một số thông số, kết quả sau khi chạy thuật toán.

### A. Thử nghiệm với bộ dữ liệu chuẩn UCI

Cụ thể:
- Dữ liệu hoa, Iris: số chiều 4, số cụm 3, số mẫu 150
- Dữ liệu rượu vang, Wine: số chiều 13, số cụm 3, số mẫu 178

Hai thuật toán HAmFCM và FCM cũng sẽ được chạy lần lượt 20 lần trên mỗi bộ dữ liệu. Tham số cho HAmFCM vẫn là dải giá trị cho trọng số mũ, còn tham số cho FCM là một số giá trị trọng số mũ cụ thể. Các con số cụ thể có thể thay đổi theo từng bộ dữ liệu, sẽ được ghi cụ thể ở phần thông báo kết quả. Đối với dữ liệu thực tế, ta chỉ quan tâm đến trường hợp tốt nhất, do đó phần kết quả độ chính xác cũng chỉ nêu các thông số cho trường hợp tốt nhất. Tập mờ trọng số mũ ngôn ngữ tìm được sẽ được chỉ ra cho mỗi trường hợp phân cụm. Bên cạnh đó, tham số mờ của HA tìm được cũng được nêu ra để tiện theo dõi, bao gồm:

$$f_m(small), f_m(big), \mu(less), \mu(possibly), \mu(more), \mu(very)$$

Để tham khảo thêm, ta cũng đưa ra kết quả thử nghiệm được công bố của các thuật toán cải tiến đã có là FCMT2I và FCMT2G. Các kết quả này có thể kiểm chứng trong tài liệu tham khảo [2] và [8]. Ngoài ra, một số thông tin không được công bố của các thuật toán tham khảo (thời gian chạy, số vòng lặp) sẽ được điền kí hiệu N/A.

**Kết quả:**

Bộ dữ liệu Iris:

**Bảng 1.** Kết quả cho bộ dữ liệu Iris

| Tiêu chí | HAmFCM | FCM | | | FCMT2I | FCMT2G |
|---|---|---|---|---|---|---|
| | 1.5-20 | 1.5 | 10 | 20 | | |
| Độ chính xác (%) | **96.67** | 96 | 96 | 60 | 95.73 | 96.13 |
| Số vòng lặp trung bình | 15 | 9 | 24 | 1 | N/A | N/A |
| Thời gian trung bình (giây) | 7.5 | 1.4 | 3.8 | 0.2 | N/A | N/A |

Tham số mờ:

- $f_m(small) = 0.5556, f_m(big) = 0.4444,$
- $\mu(less) = 0.4466, \mu(possibly) = 0.1089\ \mu(more) = 0.2291, \mu(very) = 0.2154$

Tập mờ trọng số mũ:

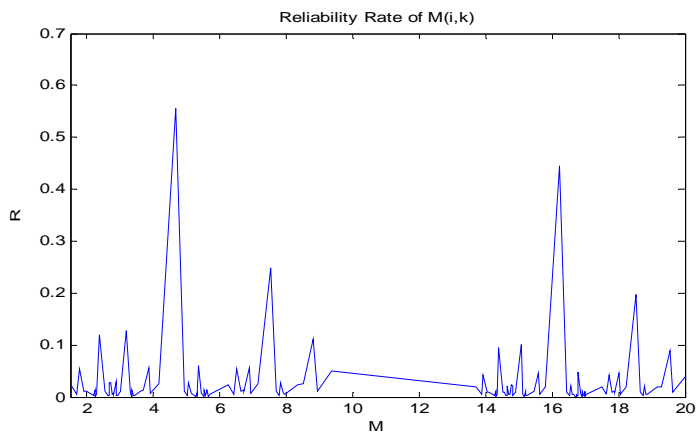

**Hình 2.** Tập mờ trọng số mũ cho bộ dữ liệu Iris

Bộ dữ liệu Wine:

**Bảng 2.** Kết quả cho bộ dữ liệu Wine

| Tiêu chí | HAmFCM | FCM | | | FCMT2I | FCMT2G |
|---|---|---|---|---|---|---|
| | 1.1-40 | 1.1 | 10 | 40 | | |
| Độ chính xác | **96.62** | 95.51 | 26.96 | 26.96 | 95.24 | 94.66 |
| Số vòng lặp trung bình | 9 | 10 | 14 | 13 | N/A | N/A |
| Thời gian trung bình | 5.1 | 1.6 | 2.2 | 2.0 | N/A | N/A |

Tham số mờ:

- $f_m(small) = 0.4386, f_m(big) = 0.5614$
- $\mu(less) = 0.2924, \mu(possibly) = 0.1462, \mu(more) = 0.2121, \mu(very) = 0.3493$

Tập mờ trọng số mũ:

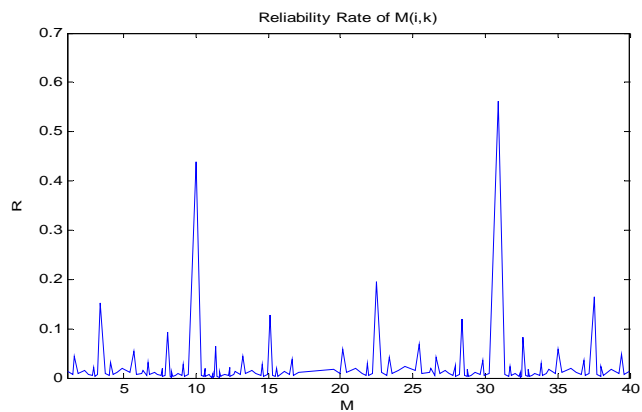

**Hình 3.** Tập mờ trọng số mũ cho bộ dữ liệu Wine

Nhận xét:

- Trong cả hai trường hợp, HAmFCM đều cho tỉ lệ phân cụm chính xác nhỉnh hơn FCM và tương đương với các thuật toán cải tiến đã có (chênh lệch từ 1-2%)
- So với 2 thuật toán cải tiến khác thì HAmFCM cũng cho thấy kết quả tốt tương đương và có phần nhỉnh hơn
- Thời gian trung bình của HAmFCM như thường lệ vẫn lâu hơn của FCM
- Số vòng lặp trung bình của HAmFCM vẫn tương đương FCM

### B. Thử nghiệm ứng dụng phân vùng ảnh màu

Nếu coi mỗi pixel trên một bức ảnh là một phần tử có giá trị (R,G,B) là giá trị màu của pixel. Ta có thể chuyển bài toán phân vùng ảnh màu, thành bài toán phân cụm như sau:

- Số phần tử là số pixel trên bức ảnh
- Giá trị của mỗi phần tử là (R,G,B): 3 chiều
- Số cụm là số vùng ảnh ta muốn phân hoạch

Ta sẽ lần lượt chạy hai thuật toán trên 3 ảnh khác nhau, tương ứng với số cụm cần tìm trên ảnh được chọn lần lượt là 2,3 và 4. Mỗi bức ảnh, từng thuật toán sẽ chạy 10 lần và chọn ra bức ảnh sau khi phân vùng có hình ảnh thị giác hợp lý nhất. Do một bức ảnh phổ thông thường có hàng trăm ngàn, thậm chí hàng triệu pixel, nếu chạy trực tiếp trên bức ảnh đó sẽ cho thời gian xử lý rất lâu, ngay cả với thuật toán FCM. Do đó, trước khi áp dụng thuật toán phân cụm vào ảnh, ảnh đã được thu nhỏ đến kích cỡ chuẩn là $48 \times 48$. Như thường lệ, tham số cho HAmFCM vẫn là dải giá trị cho trọng số mũ, còn tham số cho FCM là một số giá trị trọng số mũ cụ thể.

Tham số sử dụng trong thử nghiệm này là:
- HAmFCM: $m \in [2\ 40]$
- FCM: $m = 2, 4, 10, 15, 20, 40$. Sau khi chạy FCM cho cả 3 trường hợp chọn tham số cho kết quả tốt nhất

Ảnh kết quả được sắp xếp theo trình tự: ảnh gốc - ảnh phân vùng bởi HAmFCM - ảnh phân vùng bởi FCM.

*Kết quả cho ảnh 2 phân vùng*:

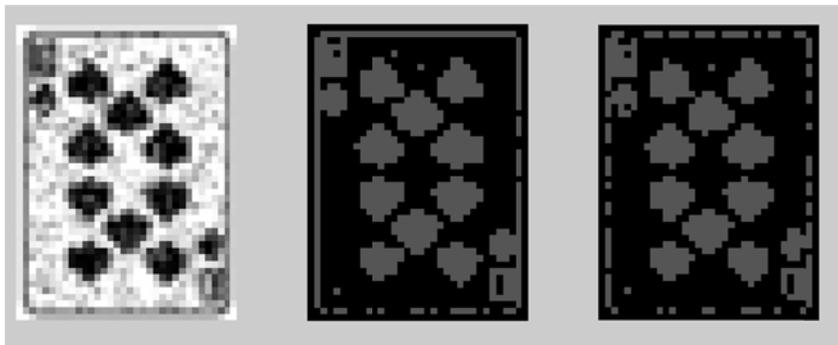

**Hình 4.** Kết quả phân cụm ảnh 2 phân vùng

Nhận xét:

- Ảnh gốc có 2 vùng cần phân biệt là phần nền trắng và các kí hiệu màu đen trên quân bài, do ảnh kích thước nhỏ nên độ nhiễu là cao
- Ảnh phân vùng do HAmFCM và FCM nói chung tương đối giống nhau trong trường hợp này
- Tuy nhiên, để ý kĩ sẽ thấy HAmFCM giữ được đường viền tương đối liền mạch ở mép trên và hai bên của ảnh gốc, trong khi FCM thất bại
- Tham số tốt nhất của FCM trong trường hợp này là $m = 2$

*Kết quả cho ảnh 3 phân vùng*:

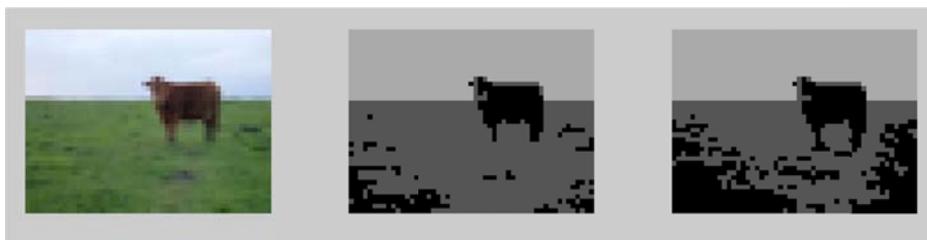

**Hình 5.** Kết quả phân cụm ảnh 3 phân vùng

Nhận xét:

- Ảnh gốc có 3 vùng cần phân biệt là vùng trời, vùng con bò và vùng cỏ
- HAmFCM cho kết quả phân vùng trời và vùng con bò giống với FCM
- Tuy nhiên, ở vùng cỏ, HAmFCM rõ ràng làm tốt hơn hẳn FCM khi xác định nhiều pixel thuộc vùng cỏ hơn. FCM xác định nhầm khá nhiều pixel vùng cỏ thành các pixel thuộc cụm dành cho vùng con bò, dẫn đến chất lượng ảnh phân vùng tệ hơn
- Tham số tốt nhất của FCM trong trường hợp này là $m = 10$

*Kết quả cho ảnh 4 phân vùng:*

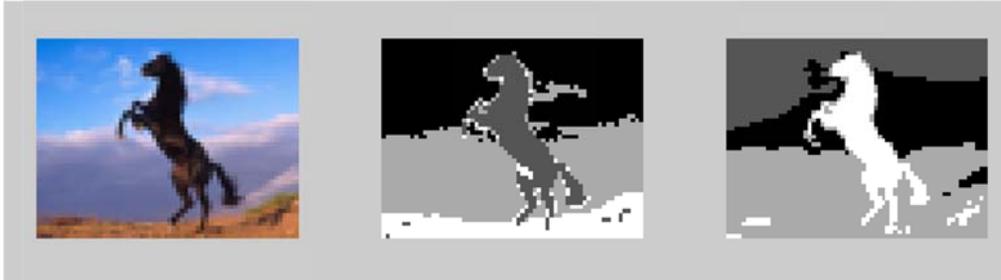

**Hình 6.** Kết quả phân cụm ảnh 4 phân vùng

Nhận xét:

- Ảnh gốc có 4 vùng cần phân biệt là vùng trời, vùng mây, vùng đất và vùng con ngựa
- HAmFCM cho kết quả phân vùng khá tốt khi chỉ ra rõ các pixel thuộc các vùng tương ứng trong khi FCM có thể nói là đã thất bại khi không phân biệt được vùng đất và vùng mây trong bức ảnh
- Tham số tốt nhất của FCM trong trường hợp này là $m = 2$

## V. KẾT LUẬN

      Bài báo đã đề xuất phương án sử dụng đại số gia tử như một mô hình mờ hóa trọng số mũ của thuật toán FCM. Kết quả thử nghiệm bước đầu cho thấy, thuật toán HAmFCM có khả năng giải quyết bài toán phân cụm tốt hơn hẳn so với thuật toán FCM trong nhiều tình huống và điều kiện đa dạng. Độ chính xác phân cụm của HamFCM có thể so sánh tương đương với các thuật toán cải tiến đã có như FCMT2I và FCMT2G.

## VI. TÀI LIỆU THAM KHẢO

# FUZZY CLUSTERING USING LINGUISTIC-VALUED EXPONENT

**Le Thai Hung, Tran Dinh Khang, Le Van Hưng**

***ABSTRACT -*** *The purpose of this paper is to study the algorithm FCM and some of its famous innovations, analyze and discover the method of applying hedge algebra-a theory that uses algebra to represent linguistic-valued variables, to FCM. Then, this thesis will propose a new FCM-based algorithm which uses hedge algebra to model FCM's exponent parameter. Finally, the design, analysis and implementation of the new algorithm as well some experimental results will be presented to prove my algorithm's capacity of solving clustering problems in practice.*